\documentclass[final]{cvpr}

\usepackage{times}
\usepackage{epsfig}
\usepackage{graphicx}
\usepackage{amsmath}
\usepackage{amssymb}

\usepackage[table]{xcolor}
\usepackage{url}
\usepackage{algorithm}
\usepackage[noend]{algpseudocode}
\usepackage{xspace}
\usepackage{booktabs}
\usepackage{multirow}
\usepackage{siunitx}

\usepackage{caption}
\usepackage{subcaption}

\definecolor{teal}{HTML}{008080}
\definecolor{indigo}{HTML}{4b0082}
\definecolor{olive}{HTML}{808000}

\usepackage[pagebackref=true,breaklinks=true,colorlinks,bookmarks=false]{hyperref}

\begin{document}

\title{On Negative Sampling for Audio-Visual Contrastive Learning from Movies}

\author{Mahdi M.~Kalayeh \enspace Shervin Ardeshir \enspace Lingyi Liu\thanks{Equal Contribution.} \enspace Nagendra Kamath\footnotemark[1]\\
Netflix Research\\
{\tt\small \{mkalayeh,shervina,nkamath,lliu\}@netflix.com}
\and
Ashok Chandrashekar\thanks{The author was with Netflix when this work started.}\\
WarnerMedia\\
{\tt\small ashok.chandrashekar@warnermedia.com}
}

\maketitle

\begin{abstract}
    The abundance and ease of utilizing sound, along with the fact that auditory clues reveal a plethora of information about what happens in a scene, make the audio-visual space an intuitive choice for representation learning. In this paper, we explore the efficacy of audio-visual self-supervised learning from uncurated long-form content i.e movies. Studying its differences with conventional short-form content, we identify a non-i.i.d distribution of data, driven by the nature of movies. Specifically, we find long-form content to naturally contain a diverse set of semantic concepts (semantic diversity), where a large portion of them, such as main characters and environments often reappear frequently throughout the movie (reoccurring semantic concepts). In addition, movies often contain content-exclusive artistic artifacts, such as color palettes or thematic music, which are strong signals for uniquely distinguishing a movie (non-semantic consistency). Capitalizing on these observations, we comprehensively study the effect of emphasizing within-movie negative sampling in a contrastive learning setup. Our view is different from those of prior works who consider within-video positive sampling, inspired by the notion of semantic persistency over time, and operate in a short-video regime. Our empirical findings suggest that, with certain modifications, training on uncurated long-form videos yields representations which transfer competitively with the state-of-the-art to a variety of action recognition and audio classification tasks.
\end{abstract}

\section{Introduction}\label{sec:introduction}
Recently, there has been tremendous progress in self-supervised learning from still images, where the standard supervised training has been outperformed in a variety of image-related tasks \cite{chen2020simple,chen2020big,he2020momentum,misra2020self}. The appeal of detaching representation learning from human annotations is rooted not only in the non-trivial challenges of scaling-up the labeling process, but also in the ill-defined task of determining a proper taxonomy with generalization power and transferability. Both challenges only exacerbate as we move from images to videos, where the notion of time is involved and the complexity of visual concepts increases. Simply considering the number of training instances, or even the cardinality of the label set is not sufficient to conclude if one large-scale supervised dataset is more suitable than another for transfer learning in video classification tasks \cite{kataoka2020would}. That is, the abundance of attention which video self-supervised learning has lately received is only to be expected. While many research efforts in this area extend the contributions made initially in the image domain to the video domain, others, including our work, have explored harnessing additional modalities such as audio or text for multi-modal self-supervised learning \cite{alayrac2020self,alwassel2019self,arandjelovic2017look,korbar2018cooperative, miech2020end,morgado_avid_cma,owens2016ambient,owens2018audio,patrick2020multi}. 

The quality of learned representations though, evaluated by transfer learning on downstream tasks, is heavily influenced by the size and taxonomy of the pretraining datasets \cite{alayrac2020self,alwassel2019self,patrick2020multi}. Different implicit and explicit assumptions on the data, often lead to different architecture designs, loss functions, sampling strategies, and notions of similarity, which may yield drastically different outcomes depending on the nature of the input data. A clear example of such findings is discussed in \cite{NEURIPS2020_22f791da}, in which the authors show that changing the pretraining data from the object-centric images of Imagenet\cite{deng2009imagenet} to more scene-centric ones in MSCOCO\cite{lin2014microsoft} could significantly impact the representations that are learned in a self-supervised contrastive learning setup. 
\begin{figure*}[ht!]
    \centering
    \includegraphics[width=\linewidth]{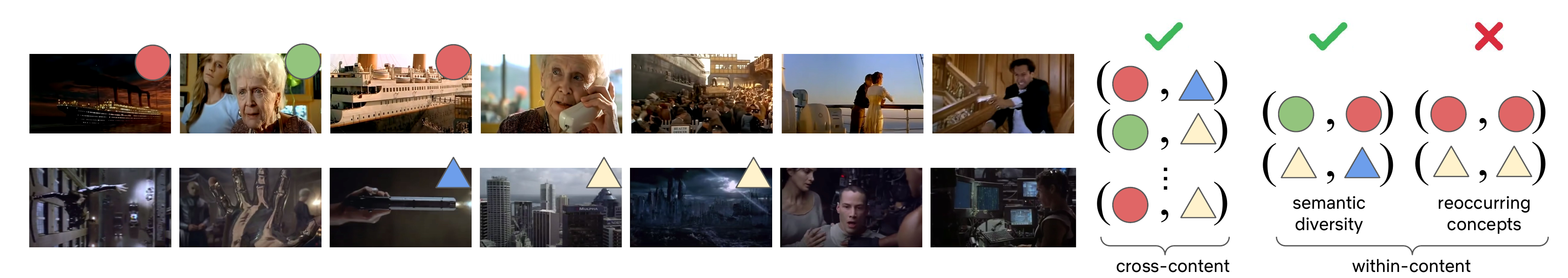}
    \caption{The non-i.i.d nature of the data distribution in the long-form content: Left illustrates sampled frames from two different movies, one in each row, where we can observe \textit{non-semantic consistency}, in form of color pallet driven artifacts (bottom row is generally darker), across different clips of each movie. Due to \textit{semantic diversity}, negative sampling from within a movie is safe as it would mostly result in semantically non-correspondent pairs of clips. However, because of \textit{reoccurring semantic concepts}, there still exists a possibility of constructing a negative pair from semantically similar clips which will not be ideal.}
    \label{fig:teaser}
\end{figure*}
Similarly, when it comes to learning from videos, self-supervised learning literature mostly uses \textit{large} catalogues of \textit{short} videos \cite{kay2017kinetics, ghadiyaram2019large}. The short length of the videos, and the shear number of them has led to an underlying i.i.d assumption of data distribution, based on which many of the prior works have been developed. With the data of such nature, there is often an implicit assumption of within-video semantic consistency \cite{feichtenhofer2021large, qian2021spatiotemporal}, which is intuitive as the likelihood of a short video containing a single semantic concept or at least very coherent ones is relatively high. On that basis, prior works \cite{feichtenhofer2021large, patrick2021space, qian2021spatiotemporal} treat different clips of a given video as augmentations of the same semantic concept. Hence, minimizing a contrastive objective is set to encourage two clips that are sampled from the same video to become more similar in the latent embedding space, while repelling pairs where clips come from two different source video instances. In this work, we argue that such an assumption is not universal, and in fact is sub-optimal when learning from long-form content like movies. In the following, we identify three main characteristics for the distribution of clips that are derived from a collection of long-form contents.

\textbf{Semantic Diversity.} Long-form content often\footnote{For instance in stand-up comedy, visuals are temporally persistent.} contains a diverse set of semantic concepts, such as characters, actions, scenes and environments. Thus, different from the short-video regime, two random clips from the same long-form content are more likely to be semantically dissimilar. This characteristic encourages within-content negative sampling, as is shown in the Figure \ref{fig:teaser}. 

\textbf{Non-Semantic Consistency.} Movies usually have underlying attributes such as color palettes, thematic background music, and other artistic patterns, some also as a result of post production. These \textit{artifacts}, often consistent throughout the content, are independent of the audiovisual semantics that are being depicted. We argue that considering all clips of a long-form video to be semantically correspondent, as is practiced in the prior works like \cite{feichtenhofer2021large, patrick2021space, qian2021spatiotemporal}, could lead to the model relying on such semantically irrelevant artifacts and ignoring the semantics of the content. Given its independence from audiovisual semantics, such characteristic suggests constructing negative pairs from a single long-form content \textit{i.e} within-content negative sampling. Figure \ref{fig:teaser} illustrates how the difference in the color palettes of two movies could provide a much easier route for the optimization to take, than actually learning the visual semantics when repelling a cross-content negative pair.   

\textbf{Reoccurring Semantic Concepts.} There are often concepts such as scenes, environments, and characters, which re-appear with minute variations throughout a movie. Thus, even though two random clips of the same long-form content are likely semantically dissimilar, the possibility of a semantic correspondence even between temporally distant clips still exists. This characteristic will theoretically lead to the \textit{class-collision} phenomenon \cite{arora2019theoretical} which is naturally the price of negative sampling on any unlabeled dataset. However, we argue that due to the aforementioned \textit{semantic diversity}, its likelihood is relatively low, because a random pair is significantly more likely to represent different concepts as semantic diversity grows.

Earlier, we alluded to the fact that the self-supervised learning literature mostly uses \textit{large} catalogues of \textit{short} videos \cite{kay2017kinetics, ghadiyaram2019large}. On top of the short duration, videos in these datasets (with a few exceptions) are well-trimmed, and semantically curated. This is crucially important since the taxonomy of downstream benchmarks are often encapsulated by, or are largely related to those of the pretraining datasets. To clarify our terminology, as \textit{curation} could be an overloaded term, we distinguish between \textit{artistic curation}, and \textit{semantic curation}. We argue that a dataset of movies, even though heavily produced and \textit{artistically} curated, is \textit{semantically} uncurated.  
Semantically curated data refers to the likes of \textit{supervised} large-scale action recognition and audio classification datasets such as Kinetics \cite{carreira2017quo}, IG-Kinetics \cite{ghadiyaram2019large}, AudioSet \cite{gemmeke2017audio}, and YouTube-8M \cite{abu2016youtube}. While the human-annotated labels are not accessed for self-supervised pretraining, videos being trimmed and from a label set of limited cardinality with biased sampling distribution \cite{alwassel2019self} acts as an implicit supervision. On the other hand, a semantically \textit{uncurated} data refers to likes of IG-Random\cite{alwassel2019self}, simply a body of unlabeled videos collected blindly with none of the aforementioned careful human-involvements. We argue that using closed-set semantically curated datasets of well-trimmed videos, is analogous to using clean object-centric images (as mentioned in \cite{NEURIPS2020_22f791da}), and could overestimate the efficacy of a self-supervised pretraining regime, especially if downstream evaluations focus on benchmarks with similar characteristics. 

Our work aims at comprehensively exploring the efficacy of learning from movies, as a long-form and semantically uncurated data, for audio-visual self-supervised learning. The three characteristics mentioned earlier, suggest exploring within-content negative sampling, with the possibility of diminishing returns past a certain level of emphasis. We explore such hypothesis, and experiment with the extent to which negative sampling could be helpful in this context.          


\section{Related Work}\label{sec:related-work}
Self-supervised learning techniques define \textit{pretext} tasks, mostly inspired by the natural structures in the data, in order to generate supervisory signals for training. Despite the plethora of proposed \textit{pretext} tasks in the literature, these approaches can be coarsely divided into two groups, namely \textit{pretext learning}, and \textit{pretext-invariant} methods. Approaches which fall in the former bucket, usually apply a form of transform, randomly drawn from a parametric family, to the input data then optimize for predicting the parameters of the chosen transformation. Predicting the relative position of image patches \cite{doersch2015unsupervised}, solving jigsaw puzzles \cite{noroozi2016unsupervised}, estimating artificial rotations \cite{gidaris2018unsupervised}, colorization \cite{zhang2016colorful}, context encoders learned through inpainting \cite{pathak2016context}, and learning by counting scale and split invariant visual primitives \cite{noroozi2017representation}, are among many methods which belong to this category. Similar techniques have been extended from images to videos \cite{fernando2017self,kim2019self,lai2019self,lee2017unsupervised,misra2016shuffle,vondrick2018tracking,wang2019self,xu2019self}, where in addition to the spatial context, the temporal domain, and the arrow of time have been heavily exploited. In contrast, \textit{pretext-invariant} methods \cite{bachman2019learning,chen2020simple,chen2020big,he2020momentum,hjelm2018learning,henaff2019data,misra2020self,oord2018representation,patrick2020multi,tian2019contrastive} are built on the concept of maximizing mutual information across augmented versions of a single instance, and are mostly formulated as contrastive learning. In other words, a pretext is used to generate different views of a single input for which the learning algorithm aims to maximize the intra-instance similarity, across variety of transformations. Our work falls within this category, however we function in a multi-modal realm employing both audio and video.

Earlier works have harnessed audio and video for representation learning through temporal synchronization \cite{korbar2018cooperative,owens2018audio}, correspondence \cite{arandjelovic2017look}, context-prediction\cite{recasens2021broaden}, and cross-modal clustering \cite{alwassel2019self, owens2016ambient}. Patrick \textit{et al.}\cite{patrick2020multi} have proposed a generalized data transformation to unify a variety of audiovisual pretext tasks through a noise contrastive formulation. This work is close to ours in the choice of objective function and data type, yet we employ no augmentation (except \textit{modality projection} in the terminology of \cite{patrick2020multi}), and solely focus on capitalizing the advantages of learning from long-form content. Morgado \textit{et al.}\cite{morgado_avid_cma} have shown that cross-modal discrimination is important for learning good audio and video representations. This was similarly pointed out earlier by \cite{alwassel2019self} in a clustering framework. Beyond that, \cite{morgado_avid_cma} generalizes the notion of instance-level positive and negative examples by exploring cross-modal agreement where multiple instances are grouped together as positives by measuring their similarity in both the video and audio feature spaces. While we also adopt a cross-modal noise contrastive estimation loss, we stick to the vanilla version, instance-level positive and negatives, and do not use any memory bank feature representations. Finally, Alayrac \textit{et al.}\cite{alayrac2020self} recently proposed a multi-modal versatile network capable of simultaneously learning from audio, video and text. Building on the intuition that different modalities are of different semantic granularity, audio and video are first compared in a fine-grained space while text is compared with the aforementioned modalities in a lower dimensional coarse-grained space. 

As mentioned earlier, prior works have been developed around large scale datasets of short videos, and on an implicit assumption of within-content semantic consistency. More specifically \cite{feichtenhofer2021large} matches a query clip, to multiple key clips in the same video, to encourages feature persistency over time, which clearly relies on the aforementioned assumption. Similarly, \cite{qian2021spatiotemporal} uses temporally consistent spatial augmentation, and  \cite{patrick2021space} proposes using spatial and temporal cropping of the same source content in the latent space to establish correspondences. On the contrary, we argue that such assumption does not hold in the long-form content regime, due to its data distribution characteristics, which were discussed earlier. To the best of our knowledge, we are the first to draw this distinction, and extensively study its implications on contrastive self-supervised learning from long-form content.  


\section{Approach}\label{sec:approach}

\textbf{Notations and Architecture.} Our pretraining dataset is denoted by $\mathcal{X}=\{\mathcal{X}_{n}| n\in[1\cdots N]\}$, where $\mathcal{X}_{n}=\{x_{n,m}| m\in[1\cdots M_{n}]\}$ contains $M_{n}$ non-overlapping audiovisual snippets which are temporally segmented from the duration of the $n^{th}$ long-form content (movie) in the dataset. Each snippet includes both audio and video modalities, formally $x_{n,m}=(a_{n,m},v_{n,m})$, where $a_{n,m}\in\mathbb{R}^{1 \times P \times Q}$ and $v_{n,m}\in\mathbb{R}^{3 \times T \times H \times W}$. $T$, $H$, and $W$ denote the number of frames, height and width of the video, while $P$, and $Q$ respectively stand for the number of mel filters, and audio frames. Video and audio are processed through 18-layers deep R(2+1)D \cite{tran2018closer} and ResNet \cite{he2016deep} architectures, respectively referred to as $f:\mathbb{R}^{3} \rightarrow \mathbb{R}^{d_{f}}$ and $g:\mathbb{R}^{1} \rightarrow \mathbb{R}^{d_{g}}$. Inspired by \cite{chen2020simple}, we use \textit{projection heads}, $h_f:\mathbb{R}^{d_{f}} \rightarrow \mathbb{R}^{d}$ and $h_g:\mathbb{R}^{d_{g}} \rightarrow \mathbb{R}^{d}$, to map corresponding representations into a common $d$-dimensional space before computing the contrastive loss. The shallow architecture of $h_f$ and $h_g$ consists of two convolution layers, separated by Batch Normalization \cite{ioffe2015batch} and ReLU \cite{nair2010rectified}, followed by global average pooling. Once self-supervised pretraining is finished, we discard the projection heads and use $f$ and $g$ for transfer learning on respective downstream tasks.

\textbf{Loss Function.} With a slight abuse of notation\footnote{$i$ enumerates elements in the minibatch.}, $\mathcal{B}=\{x_i=(a_i,v_i)|i\in[1\cdots B]\}$ represents a minibatch of size $B$, where video and audio modalities associated with the $i^{th}$ instance, $x_{i}$, are denoted by $v_i$ and $a_i$. We use $z^{i}_{v}=h_f(f(v_i))$ and $z^{i}_{a}=h_g(g(a_i))$ to represent the associated embeddings generated by projection heads, and optimize the noise contrastive estimation (NCE) loss\cite{gutmann2010noise} shown in Equation \ref{eq:loss} in order to maximize the symmetric joint probability between a corresponding audio and video. For the $i^{th}$ element in the minibatch, $(z^{i}_{v},z^{i}_{a})$ serves as the positive pair, while assuming negative pairs for both modalities, $\mathcal{N}_i=\{(z^{i}_{v},z^{j}_{a}),(z^{j}_{v},z^{i}_{a})|j\in[1\cdots B],i\ne j\}$ constitutes the set of negative pairs. 

\begin{equation}\label{eq:loss}
    \mathcal{L} = -\sum_{i=1}^{B}\log\Bigg(\frac{e^{(z^{i}_{v})^\intercal (z^{i}_{a})}}{e^{(z^{i}_{v})^\intercal (z^{i}_{a})} + \displaystyle\sum_{(z'_{v},z'_{a})\in \mathcal{N}_i}e^{(z'_{v})^\intercal (z'_{a})}}\Bigg)
\end{equation}


\textbf{Sampling Policy.} Contrastive loss function shown, in Equation \ref{eq:loss}, is computed over $B$ training instances, each in the form of an audiovisual snippet. A naive sampling policy may ignore the fact that snippets comprising the pretraining dataset are in fact temporal segments trimmed from longer-form contents, \textit{i.e.} movies. Such an assumption treats our training data as independent and identically distributed random variables from $\bigcup_{n=1}^{N}\mathcal{X}_{n}$, which constitutes the default sampling policy that is commonly used in the general deep learning literature. However, as detailed in Section \ref{sec:introduction}, the underlying \textit{artifacts} (within-content \textit{non-semantic consistency}), in addition to commonalities and correlations along the temporal axis of a long-form content (\textit{reoccurring semantic concepts}), contribute to breaking the previously discussed i.i.d assumption on the training data. Note that, sampling from any video data is going to be non-i.i.d by nature, yet in this case, temporal correlations extend for much longer, given that video entities are hours-long movies. Thus, it is more accurate to think of $\mathcal{X}$ as having multiple underlying domains, oriented towards exclusive properties which different long-form contents are characterized by. We hypothesize that during training, model gradually discovers previously mentioned content-exclusive artifacts, and latches onto those to quickly minimize Equation \ref{eq:loss} leading to sub-optimal generalization. The reason being $B\ll N$, hence for $n\sim\mathbb{U}(1,N)$ and $m\ne m'$, $\mathsf{P}(x_{n,m}\in \mathcal{B}\land x_{n,m'}\in \mathcal{B})$ is very low. In other words, if a naive random sampling policy is adopted, the set of negative pairs in Equation \ref{eq:loss} would mainly include audio-video pairs from two different movies. As shown in Figure \ref{fig:teaser}, this results in easy cross-content negatives. 

In order to quantitatively assess our hypothesis, we first formulate, in Equation \ref{eq:similarity-dist}, the space of similarities associated with negative pairs. Then, the relative entropy, $\mathsf{KL}(\mathcal{S}\parallel \mathcal{D})$, where $\mathsf{KL}$ denotes Kullback–Leibler divergence, measures how differently the within-content negative pairs ($\mathcal{S}$) behave from the cross-content ones ($\mathcal{D}$). We will refer to $\mathsf{KL}(\mathcal{S}\parallel \mathcal{D})$ as the \textit{discrepancy measure}.

\begin{equation}\label{eq:similarity-dist}
    (z^{n,m}_{v})^\intercal (z^{n',m'}_{a}) \sim
    \begin{cases}
      \mathcal{S}, & \text{if}\ n=n' \land m\neq m'\\
      \mathcal{D}, & \text{if}\ n\neq n' \land \forall (m,m')\\
    \end{cases}
\end{equation}

While the i.i.d assumption suggests that $\mathcal{S}$ and $\mathcal{D}$ should be similar, as we empirically illustrate later, the \textit{discrepancy measure} is in fact rather large, and grows as the self-supervised pretraining progresses, due to an abundance of easy cross-content negative pairs, eventually leading to inferior representations. To ameliorate such optimization challenge, we take a simple alternative approach that emphasizes on the within-content negative pairs by dividing the minibatch budget of $B$, across $B/k$ randomly chosen long-form contents, where we sample $k$ snippets from each. It is worth reiterating that the prior works \cite{feichtenhofer2021large, patrick2021space, qian2021spatiotemporal} encourage temporally distant segments of the same video to be similar (positive pair) in the latent embedding space. In contrast, we treat such instances as a negative pair and aim for the optimization to push them apart from one another. 

We first uniformly sample a long-form content, $n\sim\mathbb{U}(1,N)$, and then draw $k$ distinct snippets from $\mathcal{X}_{n}$, creating $\{x_{n,m}| m\in\mathcal{M}_{n}\}$, where $\mathcal{M}_{n}\subset[1\cdots M_{n}]$ and $|\mathcal{M}_{n}|=k$. This ensures that for $x_i\in\mathcal{B}$, $\mathcal{N}_i$ always includes $2k-2$ pairs sampled from the same movie to which $x_i$ belongs. By putting constraints on $\mathcal{M}_{n}$, specifically how temporally far from each other the $k$ samples are drawn, we may go one step further and to some extent control the audiovisual similarity between snippets. This serves as an additional knob to tune for hard negative sampling. The intuition is that, the larger narrative of a professionally made movie is composed of shorter units called \textit{scene}. Each scene comprises a complete event, action, or block of storytelling and normally takes place in one location and deals with one action. That is, if our samples are temporally close, it is more likely for corresponding snippets to be highly correlated and/or look/sound alike. $k \leq \texttt{max}[\mathcal{M}_{n}]-\texttt{min}[\mathcal{M}_{n}]+1 \leq w \leq M_{n}$ defines the bounds on our sampling policy, where $w$, standing for a sampling \textit{window}, determines the farthest two out of $k$ samples drawn from $\mathcal{X}_{n}$ can be. Accordingly, $w=k$ represents the case where all $k$ samples are temporally adjacent, hence the expected audiovisual similarity is maximized due to temporal continuity in content. In our preliminary studies, we observed that having such level of hard negatives, even with a small $k$, prevents proper training and results in performance degradation. On the other hand, $w=M_{n}$ indicates random sampling where no temporal constraint is imposed on $\mathcal{M}_{n}$, thus samples are less likely to be drawn from adjacent time-stamps. The rest of the spectrum provides middle grounds where two samples drawn from $\mathcal{X}_{n}$ can at most be $w+1$ snippets apart.

\section{Experiments}\label{sec:experiments}

\subsection{Experimental Setup}\label{subsec:experimental-setup}
\textbf{Datasets and Reproducibility.} We use full-length movies for self-supervised pretraining. Movies are randomly chosen from a large collection spanning over a variety of genres, namely Drama, Comedy, Action, Horror, Thriller, Sci-Fi and Romance. All audio is in English language. Our pretraining dataset consists of $\sim$3.6K movies with an average duration of 105 minutes. Given that we cannot publicly release our dataset due to copyright reasons, we acknowledge that it is not possible for other research groups to fully reproduce our results. However, we intend to make the pretrained models publicly available, and hope that the research community finds them along with the other contributions of this work of value. We would like to emphasize that similar limitations have precedents in multiple earlier works including but not limited to \cite{alwassel2019self,ghadiyaram2019large,mahajan2018exploring,sun2017revisiting,feichtenhofer2021large}. 

\textbf{Pretraining.} Unless mentioned otherwise, we use video snippets with 16 frames at 5 fps. For data augmentation, we resize the shorter side to 224 pixels, then randomly crop them into $200\times200$ pixels. As for sound, we compute mel spectrogram from the raw audio at 48K sample rate using 96 mel filters and an FFT window of 2048, while the number of samples between successive frames is set to 512. For data augmentation, we randomly drop out up to 25\% from either temporal or frequency axis of the 2-D mel spectrogram image. The dimension of audio-video joint embedding space, $d$, is set to 512. Models are trained using ADAM \cite{kingma2014adam} optimizer,  with an initial learning rate of $10^{-4}$ which linearly warms up to $0.002$ during the first epoch. We use a cosine learning rate schedule and a batch size of 384. Kernel size is 1 for convolution layers in $h_f$ and $h_g$. Unless mentioned otherwise, we pretrain for 10 epochs when reporting ablation studies and increase it to 40 for comparison with the state-of-the-art.

\textbf{Downstream Evaluation.} 
To measure the quality of the learned representations, we follow recent works \cite{alwassel2019self,patrick2020multi,morgado_avid_cma, alayrac2020self,Recasens_2021_ICCV,Patrick_2021_ICCV} and perform transfer learning on UCF101\cite{soomro2012ucf101} and HMDB51\cite{kuehne2011hmdb} for action recognition, along with ESC50\cite{piczak2015esc} for audio classification. In an effort to make our ablation studies more comprehensive, we further evaluate our models on datasets which are larger in scale, namely Kinetics-400\cite{kay2017kinetics} and VGGSound\cite{Chen20}.

For UCF101 \cite{soomro2012ucf101} and HMDB51 \cite{kuehne2011hmdb}, we use video clips that are 32 frames long at 10 fps. Unless mentioned otherwise, these clips are randomly chosen from the duration of the video instances. A scale jittering range of [224, 290] pixels is used and we randomly crop the video into $200\times200$ pixels. Furthermore, random horizontal flipping and color jittering are employed. We train for a total of 200 epochs using SGD, with an initial learning rate of $10^{-3}$ which linearly warms up to $0.2$ during the first 25 epochs. Momentum and weight decay are respectively set to 0.9 and $10^{-4}$. We use a cosine learning rate schedule and a batch size of 96. The setup is the same in both finetuning and linear evaluation regimes, except in the latter, we set the weight decay and dropout both to zero. During inference, 10 temporal clips are uniformly sampled where each is spatially cropped in 3 ways (left, center, right) resulting in a total of 30 views. We then average the model predictions across these 30 views and report top-1 classification accuracy(\text{\%}). Evaluation on Kinetics-400\cite{kay2017kinetics} follows the exact setup mentioned above, except we train for 50 epochs since the dataset is larger, with batch size and learning rate respectively set to 192 and $0.4$. 

For ESC50 \cite{piczak2015esc}, we use 3-seconds clips which are randomly chosen from the duration of the audio instances and apply time and frequency masking to spectrogram images for data augmentation. The maximum possible length of the mask is 50\% of the corresponding axis. We do not use any scale jittering or random cropping on the spectrograms. We train for a total of 200 epochs with warm up during first 25 epochs. Other optimization parameters are the same as those in the aforementioned action recognition tasks. During inference, 10 temporal clips are uniformly sampled and we average the model predictions across these 10 views and report top-1 classification accuracy(\text{\%}). Evaluation on VGGSound\cite{Chen20} follows the same setup as the one detailed for ESC50 \cite{piczak2015esc} except batch size and learning rate are respectively set to 512 and  $0.4$. We report mean average precision on VGGSound\cite{Chen20}.

\begin{figure*}[ht!]
        \centering
        \includegraphics[width=0.32\linewidth]{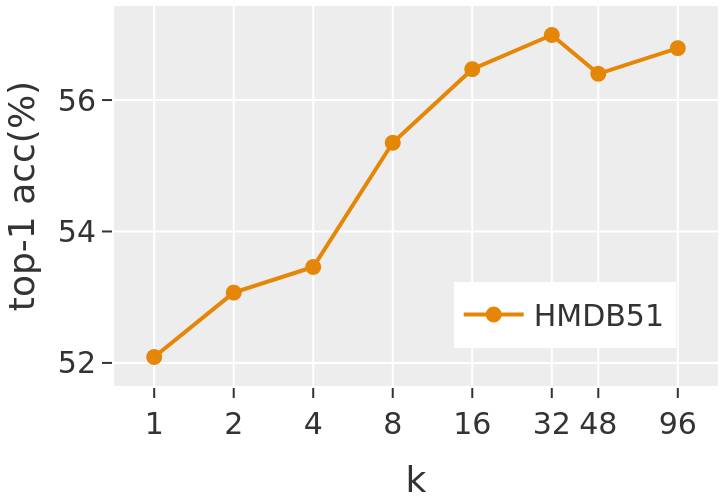}
        \includegraphics[width=0.32\linewidth]{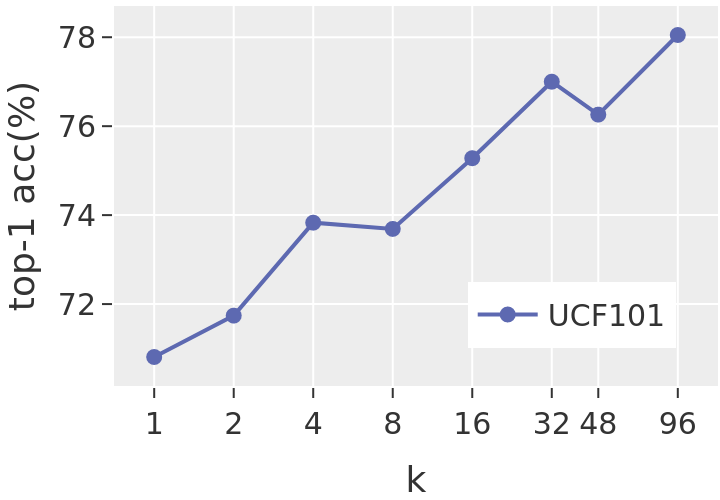}
        \includegraphics[width=0.32\linewidth]{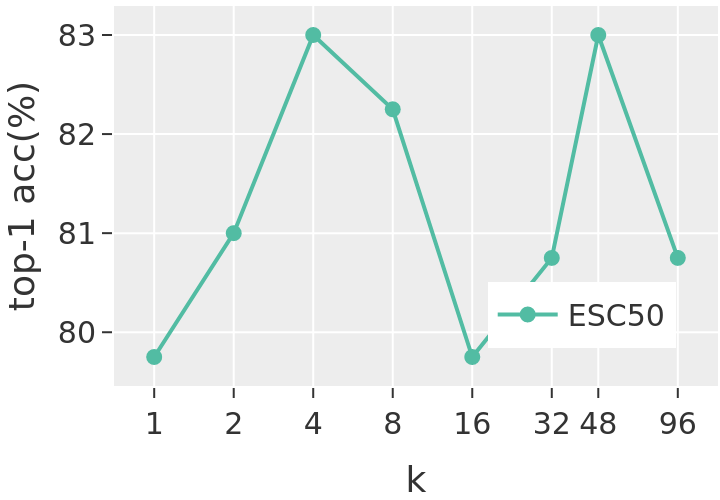}
        \includegraphics[width=0.32\linewidth]{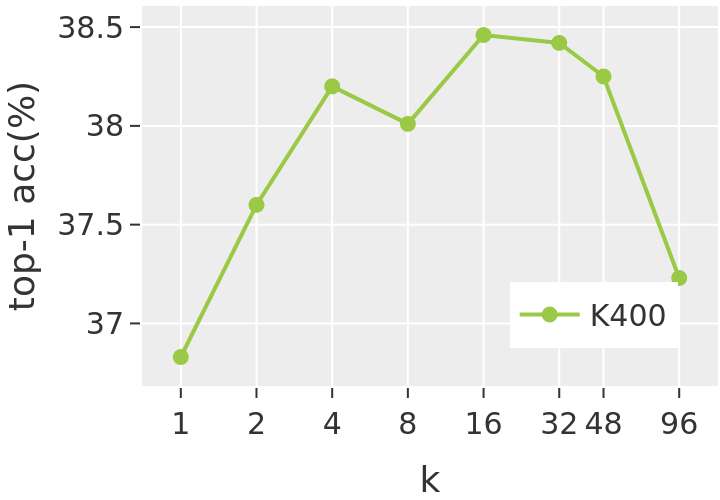}
        \includegraphics[width=0.32\linewidth]{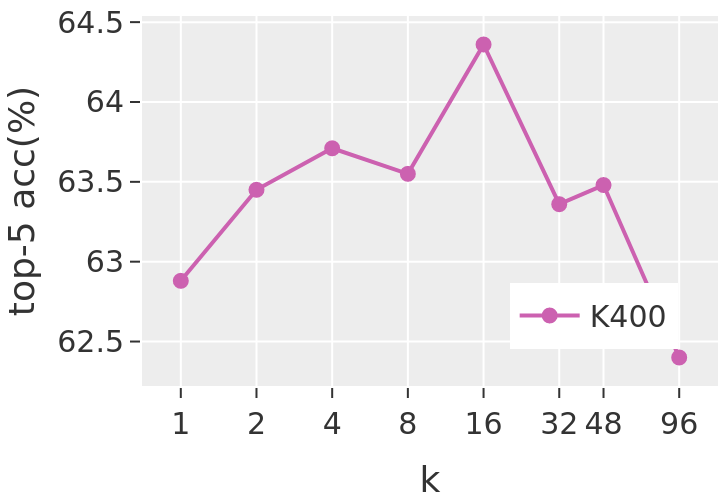}
        \includegraphics[width=0.32\linewidth]{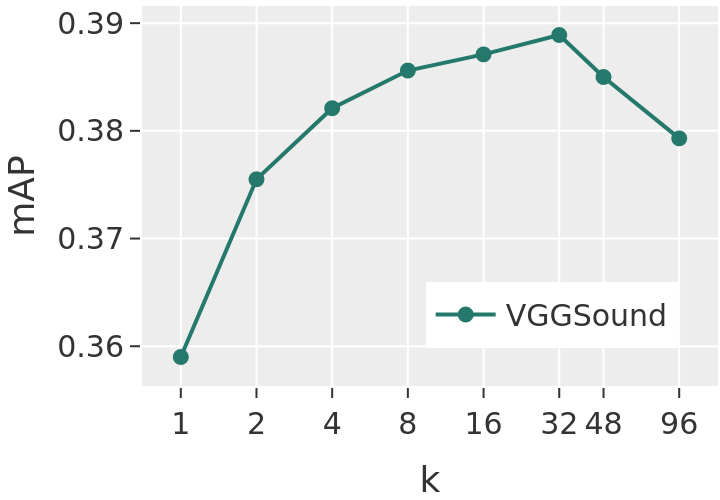}
        \caption{Effect of emphasizing on within-content negative sampling through increasing ${k}$ during pretraining. Downstream transfer learning performances are measured in a linear evaluation regime. For HMDB51\cite{kuehne2011hmdb}, UCF101\cite{soomro2012ucf101} and ESC50\cite{piczak2015esc}, numbers are reported on the split-1 of the corresponding datasets. For Kinetics-400 (K400)\cite{kay2017kinetics} and VGGsound\cite{Chen20}, we use their validation sets. Sampling window ($w$) is set to 4 times as $k$. Refer to Table \ref{tab:sampling-policy} for detailed numbers.}
        \label{fig:sampling-policy}
\end{figure*}

\subsection{Ablation Study}\label{subsec:ablation-study}
Here, we discuss multiple ablation studies to explore how emphasizing on within-content negative sampling enables better representation learning from long-form content.  

\subsubsection{Pretraining and Generalization}
\begin{figure}[h!]
        \centering
        \includegraphics[width=0.45\textwidth]{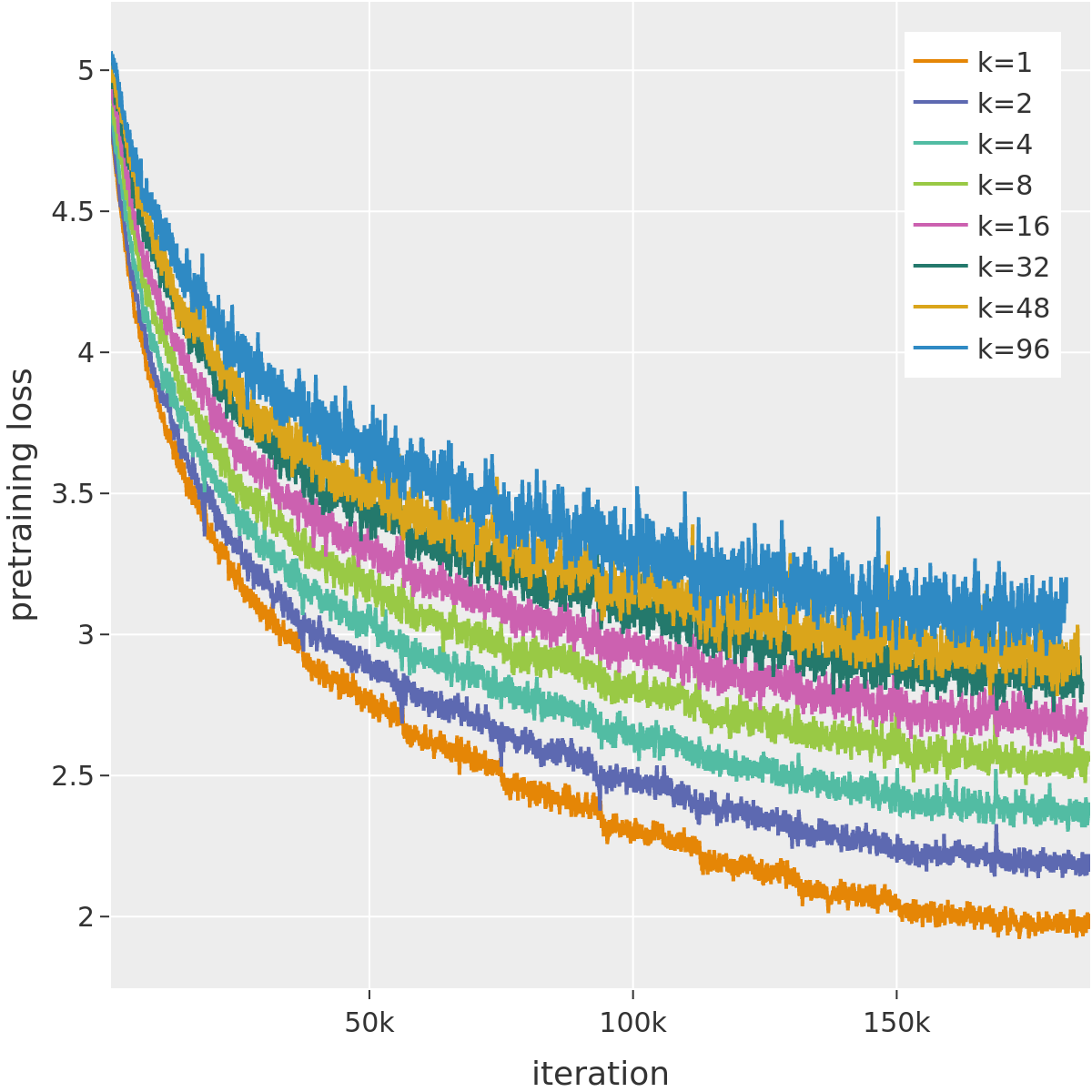}
        \caption{Effect of emphasizing on within-content negative sampling, through increasing ${k}$, on pretraining loss.}
        \label{fig:training_loss}
\end{figure}

Aligned with our expectations, Figure \ref{fig:training_loss} confirms that increasing $k$ beyond 1, which denotes the baseline i.i.d assumption, results in harder pretraining objectives as more within-content negative samples are contributing to the denominator of the Equation \ref{eq:loss}. In other words the distribution of instances in the minibatch $\mathcal{B}$ shifts from a regime of \textit{single\footnote{in expectation} clip from many videos} to \textit{many clips from few videos}. Meanwhile, increasing the difficulty of the self-supervised pretext task leads to better downstream performance. Figure \ref{fig:sampling-policy} illustrates the transfer learning results, in a linear evaluation regime on different downstream tasks, where $k>1$ clearly yields large performance improvements. With an effective batch size of 96\footnote{distributed training on 4 GPUs}, spanning $k$ across the full spectrum allows us to study how self-supervised pretraining is influenced by different amounts of video-level diversity. Recall that, according to the sampling policy, described in Section \ref{sec:approach}, minibatch $\mathcal{B}$ should consist of $96/k$ movies from each there are $k$ instances present. In general, $k>16$ is where the incremental gains seem to diminish. On Kinetics-400\cite{kay2017kinetics} in particular, we see a sharp drop when $k=96$. As was alluded to earlier, in such a setting the video-level diversity vanishes since the minibatch is comprised solely of instances that belong to a single movie. Table \ref{tab:sampling-policy} shows the detailed numbers for downstream evaluations shown in Figure \ref{fig:sampling-policy}.

\begin{table}
  \small
  \centering
  \begin{tabular}{lccccc}
	\toprule
    $k$ & HMDB51 & UCF101 & ESC50 & K400 & VGGSound\\
    \midrule
    1 & 52.09 & 70.81 & 79.75 & 36.83 & 0.3590\\
    \midrule
    2 & 53.07 & 71.74 & 81.00 & 37.60 & 0.3755\\
    4 & 53.46 & 73.83 & \textbf{83.00} & 38.20 & 0.3821\\
    8 & 55.35 & 73.69 & 82.25 & 38.01 & 0.3856\\
    16 & 56.47 & 75.28 & 79.75 & \textbf{38.46} & 0.3871\\
    32 & \textbf{56.99} & 77.00 & 80.75 & 38.42 & \textbf{0.3889}\\
    48 & 56.40 & 76.26 & \textbf{83.00} & 38.25 & 0.3850\\
    96 & 56.79 & \textbf{78.05} & 80.75 & 37.23 & 0.3793\\
    \midrule
    $\Delta$ & +4.90 & +7.24 & +3.25 & +1.63 & +0.0399\\
    \bottomrule    
  \end{tabular}
  \caption{
  Effect of emphasizing on within-content negative sampling through increasing ${k}$ during pretraining. Downstream transfer learning performances are measured in a linear evaluation regime. The best result in each column is denoted in bold while $\Delta$ indicates the gain over the baseline of $k=1$. We report mean average precision on VGGSound and top-1 classification accuracy(\text{\%}) on the other benchmarks.}
  \label{tab:sampling-policy}
\end{table}

\begin{figure*}[h!]
    \centering
    \begin{subfigure}[b]{0.32\textwidth}
        \centering
        \includegraphics[width=1\linewidth]{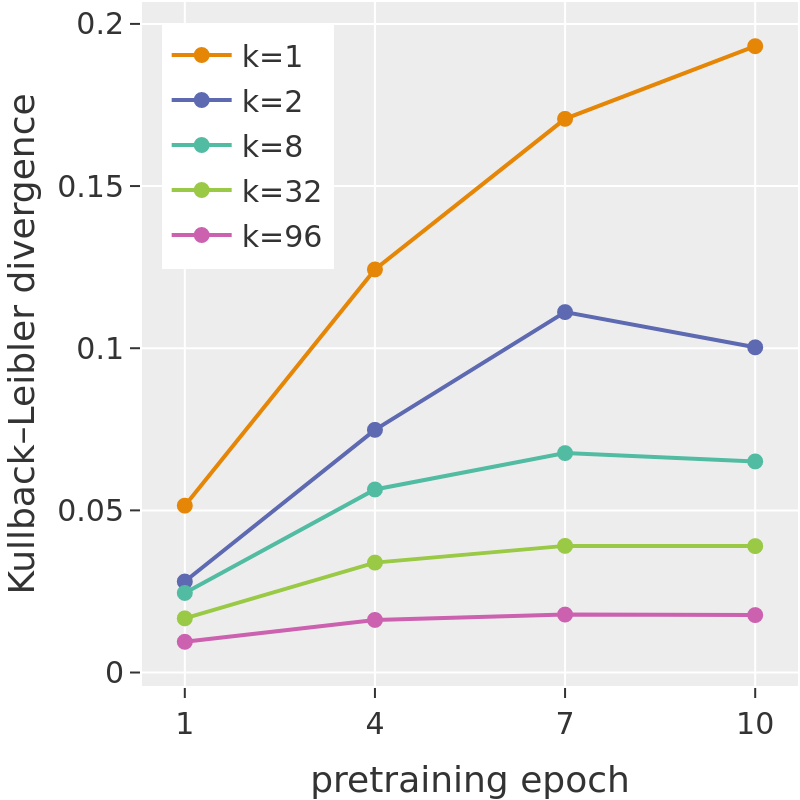}
    \end{subfigure}
    \begin{subfigure}[b]{0.32\textwidth}
        \centering
        \includegraphics[width=1\linewidth]{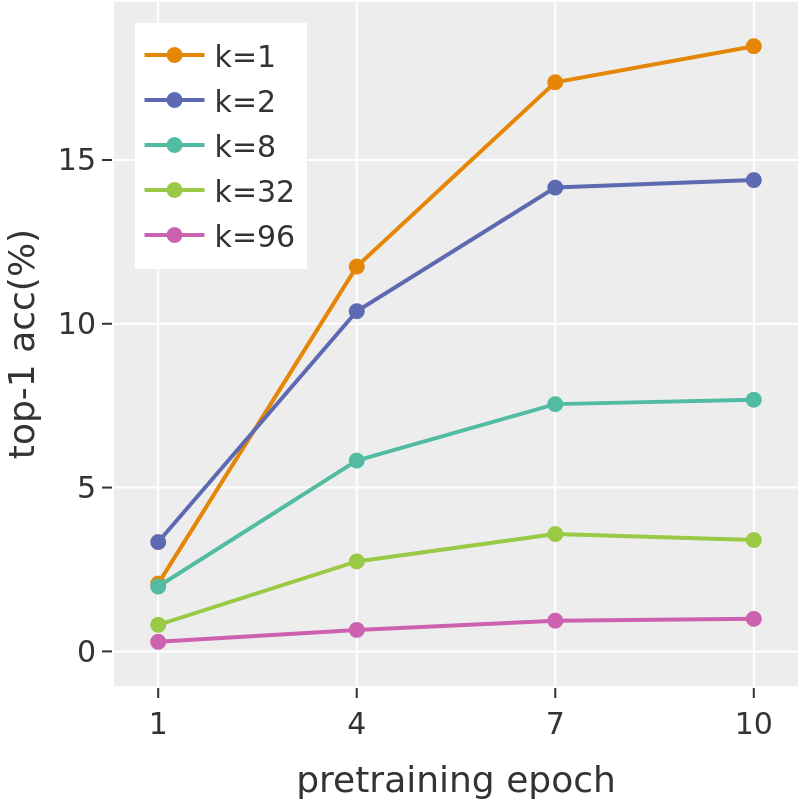}
    \end{subfigure}
    \begin{subfigure}[b]{0.32\textwidth}
        \centering
        \includegraphics[width=1\linewidth]{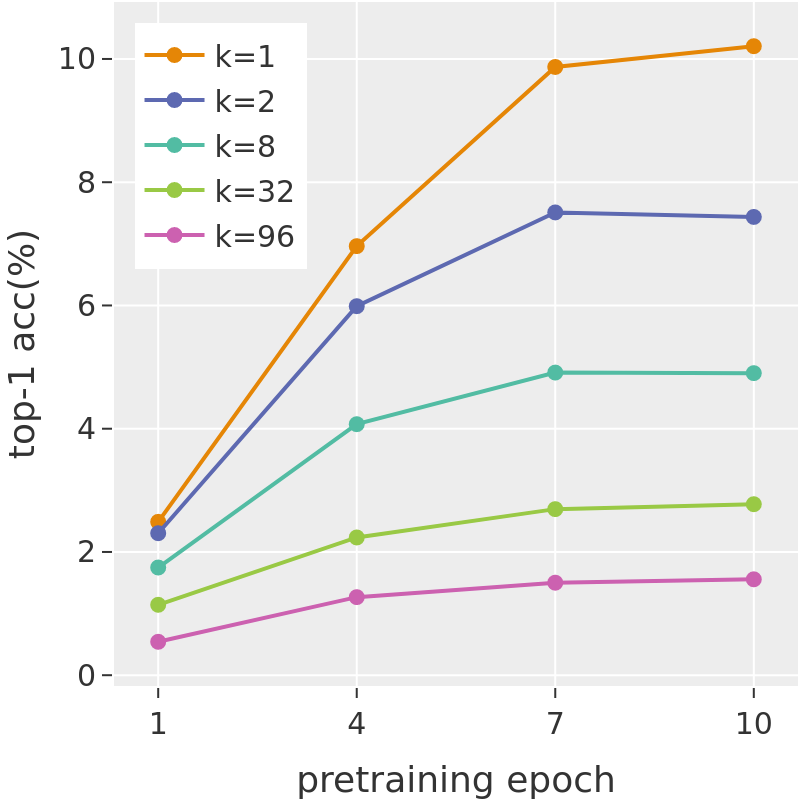}
    \end{subfigure}    
    \caption{Left: $\mathsf{KL}(\mathcal{S}\parallel \mathcal{D})$ during pretraining. Middle/Right: Video/Audio based clip-level movie classification on the held-out set.}
    \label{fig:discarding_artifacts}
\end{figure*}

\subsubsection{Discarding Content-Exclusive Artifacts} While Figure \ref{fig:sampling-policy} attests that learning relevant semantic concepts has meaningfully improved, we are interested in measuring whether the effects of content-exclusive artifacts, those which contribute to the \textit{non-semantic consistency} within each long-form content (ref. Section \ref{sec:introduction}), are also successfully discarded. Note that, these artifacts can collectively be seen as a signature that audio-visually distinguish one movie from another, yet are not semantically valuable \textit{e.g} color palette as was shown in Figure \ref{fig:teaser}. Figure \ref{fig:discarding_artifacts} shows the \textit{discrepancy measure} during pretraining, where we use the weights of the models after 1, 4, 7 and 10 epochs of pretraining to extract (offline) the final embeddings from the training data and empirically compute the symmetric Kullback–Leibler divergence between $\mathcal{S}$ and $\mathcal{D}$. Here, we can make a few observations. \textbf{First}, for the baseline $k=1$, $\mathsf{KL}(\mathcal{S}\parallel \mathcal{D})$ grows quickly as pretraining progresses, indicating that the final audio and video embeddings not only have largely preserved the the artifacts, but also the optimization increasingly exploits them instead of the semantically relevant attributes, when repelling a mismatched (negative) audio-video pair. \textbf{Second}, as we emphasize more on employing within-content negative sampling through increasing $k$, the \textit{discrepancy measure} significantly decreases, demonstrating that the contrastive loss is in fact reducing its reliance on content-exclusive artifacts. Note that such change of mechanism helps us learn better representations as was shown earlier in Figure \ref{fig:sampling-policy}. In particular, given a fixed minibatch budget, a larger $k$ favors more training instances to be sampled from fewer number of movies. That increases the portion of hard\footnote{when a mismatched pair belong to the same movie} negative pairs, thus encourages the contrastive loss to more aggressively push away pairs from the same movie, ultimately leading to less of the content-exclusive artifacts being preserved in the embedding space. \textbf{Third}, it is worth emphasizing how larger $k$, even as training progresses, prevents $\mathsf{KL}(\mathcal{S}\parallel \mathcal{D})$ from growing to the point that for $k=32$ and $k=96$, the associated curves rather plateau after the 4$^{th}$ pretraining epoch, confirming that large $k$ encourages an embedding space where the boundary between different movies is very much blurred.

Another lens through which we can measure how effective the content-exclusive artifacts have been removed is a simple clip-level movie classification task. Recall from Section \ref{sec:approach} that, $\mathcal{X}_{n}=\{x_{n,m}| m\in[1\cdots M_{n}]\}$ contains $M_{n}$ non-overlapping audiovisual snippets which are temporally segmented, while maintaining the chronological order, from the duration of the $n^{th}$ long-form content (movie) in our pretraining dataset. Similar to the previous study, we use the weights of different models after 1, 4, 7 and 10 epochs of pretraining to extract $z^{n,m}_{v}$ and $z^{n,m}_{a}$. We divide each movie into half, where $\bigcup_{n=1}^{N}\{(z_{n,m},n)| m\in[1\cdots M_{n}/2]\}$ and $\bigcup_{n=1}^{N}\{(z_{n,m},n)| m\in(M_{n}/2\cdots M_{n}]\}$ respectively construct our train and test splits\footnote{We intentionally used the first half for train and second half for test instead of a random selection as that would likely divide temporally close snippets, which sound and look similar, between train and test, effectively leading to a train-test leakage.}. Note that, here the label assigned to each train/test instance is the index of the movie from which it was segmented. Our hypothesis is as follows: by simply looking at a $d$-dimensional representation that corresponds to a $\sim$3 second snippet, we should not be able to predict from which hours-long movie the snippet was originally segmented unless there are certain movie-exclusive artifacts present in the learned representations. The more successful we are in discarding content-exclusive artifacts the harder this proxy task should become. Figure \ref{fig:discarding_artifacts} shows the top-1 classification accuracy of a Linear SVM for the task, using video and audio embeddings. We can see that the accuracy on the held-out test set significantly reduces as $k$ increases. This is inline with our intuition that larger $k$, by emphasizing on within-content negative sampling, effectively discards the movie-exclusive artifacts. On the other hand, in the regime of $k=1$, not only the artifact characteristics are preserved in the end representations but also seem to become more prominent as pretraining progresses. This observation can be drawn from how top-1 accuracy for $k=1$ significantly improves as we use representations that are extracted from models at later pretraining epochs, while that effect is pretty much non-existent for large $k$ values. Given these ablation studies, we conclude that emphasizing on within-content negative sampling results in learned representations that transfer better, are more semantically homogeneous, and carry significantly less non-semantic artifacts. 

\begin{table}
  \small
  \centering
  \begin{tabular}{llcccc}
	\toprule
	\multicolumn{6}{c}{protocol: finetuning on split-1}\\
	\midrule
    $k$ & $w$ & epochs & HMDB51 & UCF101 & ESC50\\
    \midrule
    16 & $M_n$ & 20 & 69.86 & 89.13 & 88.50\\
    16 & 256 & 20 & 70.58 & 89.79 & 87.25\\
    16 & 128 & 20 & 70.26 & 89.55 & 86.25\\
    16 & 64 & 20 & 70.52 & 89.55 & 89.25\\
    \midrule
    2 & 8 & 10 & 67.25 & 87.81 & 87.25\\
    2 & 32 & 10 & 66.73 & 88.12 & 89.00\\
    2 & 128 & 10 & 66.86 & 87.73 & 88.00\\
    \midrule
    16 & 256 & 10 & 68.23 & 88.23 & 89.25\\
    32 & 256 & 10 & 69.01 & 88.04 & 90.00\\
    64 & 256 & 10 & 68.56 & 87.67 & 86.00\\
    \bottomrule    
  \end{tabular}
  \caption{Effect of sampling window (${w}$) during self-supervised pretraining on transfer learning performance. ``epochs'' denotes the duration of pretraining.}
  \label{tab:sampling-window}
\end{table}


\subsubsection{Effect of Sampling Window} 
When drawing $k$ snippets from a long-form content, a smaller sampling window ($w$) produces instances that are temporally closer. Intuitively, this should increase the probability that samples look/sound very much alike (\textit{i.e.} harder negative pairs). Therefore, one can see the sampling window ($w$) as another knob which together with $k$ (sampling size) control the hardness of the objective function. We've previously seen that adequately increasing $k$ leads to learning representations which transfer better to different downstream tasks. Next, we are going to study if tuning for the sampling window ($w$) gives us further improvements. From Table \ref{tab:sampling-window}, we can make two observations. \textbf{First}, for a fixed $k=16$ (top block), as we gradually tighten the sampling window from $M_n$ (no temporal constraint) to 64, expecting to generate harder negative pairs, we only see a negligible additive gain in the transfer learning performance. The same behavior can be seen for $k=2$ as well. \textbf{Second}, since the difficulty of the pretext task is controlled by the interaction between $k$ and $w$, alternatively we can tune for $k$ given a fixed $w$. Table \ref{tab:sampling-window} (bottom block) seems to suggest that with an exception on ESC50\cite{piczak2015esc}, the differences are negligible. Previous ablation studies strongly indicate that large $k$ leads to learning representations that generalize better with a large margin. However, given an already large $k$, tuning for the sampling window ($w$) as is illustrated in Table \ref{tab:sampling-window}, seems to provide only a negligible incremental gain. This implies that commonalities which persist throughout the duration of a movie are sufficiently powerful signals to be exploited for effective within-content negative sampling.

\begin{table}
  \small
  \centering
  \begin{tabular}{llccc}
	\toprule
    head & $d$ & HMDB51 & UCF101 & ESC50\\
    \midrule
    conv & 512 & 68.23 & 88.23 & 89.25 \\
    mlp & 512 & 64.18 & 86.70 & 88.75 \\
    mlp & 1024 & 65.94 & 87.44 & 87.75 \\    
    \bottomrule    
  \end{tabular}
  \caption{Effect of the projection head architecture. All numbers were obtained on the split-1 of corresponding datasets through end-to-end finetuning and we report top-1 accuracy(\text{\%}). We use $k=16$ and $w=256$. Pretraining runs for 10 epochs.}
  \label{tab:preatraining-head}
\end{table}

\begin{table}
  \small
  \centering
  \begin{tabular}{cccc}
	\toprule
    epochs & HMDB51 & UCF101 & ESC50\\
    \midrule
    10 & 68.23 & 88.23 & 89.25 \\
    30 & 70.13 & 89.34 & 86.50 \\
    40 & 73.00 & 89.68 & 88.75 \\
    \bottomrule    
  \end{tabular}
  \caption{Effect of the number of self-supervised pretraining epochs. All numbers were obtained on the split-1 of corresponding datasets through end-to-end finetuning and we report top-1 accuracy(\text{\%}). We use $k=16$ and $w=256$.}
  \label{tab:preatraining-epochs}
\end{table}

\subsubsection{Projection Head and Pretraining Duration}
Table \ref{tab:preatraining-head} compares the ``conv'' based projection heads, described in Sec. \ref{sec:approach}, with the ``mlp'' alternative. The difference is that the latter collapses spatio-temporal resolution before passing embeddings to the projection heads. On the other hand, we maintain the spatio-temporal and spatial resolutions, respectively in $h_f$ and $h_g$ and perform the global average pooling just at the end. We can see that collapsing the spatio-temporal resolution hurts the performance even when the size of embedding space is twice as large. 

Finally, while by default in ablation studies, we pretrain for 10 epochs, Table \ref{tab:preatraining-epochs} shows that increasing the duration of pretraining is generally helpful, however, the gain is more pronounced on HMDB51\cite{kuehne2011hmdb}.

\begin{table}
  \small
  \centering
  \begin{tabular}{llcccc}
	\toprule
	\multicolumn{5}{c}{protocol: linear evaluation on split-1}\\
	\midrule
    method & pretraining & HMDB51 & UCF101 & ESC50\\
    \midrule
    XDC\cite{alwassel2019self} & IG-Random & 49.9 & 80.7 & 84.5\\
    XDC\cite{alwassel2019self} & IG-Kinetics & 56.0 & 85.3 & 84.3\\
    ours & movies & 63.5 & 79.8 & 82.5\\
	\toprule
	\multicolumn{5}{c}{protocol: finetuning on split-1}\\
	\midrule    
    XDC\cite{alwassel2019self} & IG-Random & 61.2 & 88.8 & 86.3\\
    XDC\cite{alwassel2019self} & IG-Kinetics & 63.1 & 91.5 & 84.8\\
    ours & movies & 73.0 & 89.7 & 88.7\\
    \bottomrule    
  \end{tabular}
  \caption{Learning from movies as a source of semantically \textit{uncurated} pretraining data.}
  \label{tab:results_vs_xdc}
\end{table}

\subsection{Pretraining on Uncurated Data} 
To the best of our knowledge, the only \textit{uncurated} dataset used in literature for audio-visual self-supervised learning is IG-Random\cite{alwassel2019self} which has 65M training videos. It is an \textit{uncurated} version of the weakly-supervised collected IG-Kinetics\cite{ghadiyaram2019large} where videos were retrieved by tags relevant to the categories in the Kinetics dataset \cite{kay2017kinetics}. Alwassel \textit{et.al} \cite{alwassel2019self} accurately argue that self-supervised pretraining on likes of IG-Kinetics\cite{ghadiyaram2019large}, and other supervised datasets for that matter, introduces additional privileges since even without using labels, training videos are still biased due to the sampling distribution (\textit{e.g.}, taxonomy of the curated dataset). In this work, from a large catalogue of movies, we've randomly selected $\sim$3.6K films, an equivalent of $~$0.7 years worth of content (30 times smaller than IG-Random\cite{alwassel2019self}), as our pretraining dataset, which as discussed in Section \ref{sec:introduction}, like IG-Random\cite{alwassel2019self} is semantically \textit{uncurated}. Table \ref{tab:results_vs_xdc} compares our approach against XDC\cite{alwassel2019self}. In the finetuning regime, our model trained on a collection of movies consistently outperforms XDC\cite{alwassel2019self} with a large margin across three different tasks. Although, the gap reduces in the linear evaluation protocol. We hypothesize that to be due to difference between distribution of pretraining movies and downstream benchmarks. Interestingly, we outperform XDC\cite{alwassel2019self} in finetuning regime even when it is trained on curated IG-Kinetics\cite{ghadiyaram2019large}, in two out of three benchmarks. It is worth noting that, unfortunately the IG-Random\cite{alwassel2019self} is not publicly available, nor is the implementation to train XDC\cite{alwassel2019self}. So, cross method-dataset experiments were not possible. However, both XDC\cite{alwassel2019self} and our work use the same backbone architectures.

\begin{table}
\small
  \centering
  \begin{tabular}{lcccc}
	\toprule
	\multicolumn{5}{c}{protocol: finetuning}\\
	\midrule
     Method & Arch. & Data & HMDB51 & UCF101\\
     \midrule
    GDT\cite{patrick2020multi} & R(2+1)D-18 & K400 & 62.3 & 90.9\\
    GDT\cite{patrick2020multi} & R(2+1)D-18 & IG-K & 72.8 & 95.2\\
    STiCA\cite{Patrick_2021_ICCV} & R(2+1)D-18 & K400 & 67.0 & 93.1\\
    AVID\cite{morgado_avid_cma} & R(2+1)D-18 & K400 & 60.8 & 87.5\\
    AVID\cite{morgado_avid_cma} & R(2+1)D-18 & AS & 64.7 & 91.5\\
    XDC\cite{alwassel2019self} & R(2+1)D-18 & IG-K & 68.9 & 95.5\\
    XDC\cite{alwassel2019self} & R(2+1)D-18 & IG-R & 66.5 & 94.6\\
    XDC\cite{alwassel2019self} & R(2+1)D-18 & K400 & 52.6 & 86.8\\
    MMV\cite{alayrac2020self} & R(2+1)D-18 & AS & 70.1 & 91.5\\
    AVTS\cite{korbar2018cooperative} & MC3-18 & AS & 61.6 & 89.0\\
    AVTS\cite{korbar2018cooperative} & MC3-18 & K400 & 56.9 & 85.8\\
    CVRL\cite{qian2021spatiotemporal} & R3D-50 & K400 & 66.7 & 92.2\\
    BraVe\cite{Recasens_2021_ICCV} & TSM-50 & AS & 75.3 & 95.6\\
    ELo\cite{piergiovanni2020evolving} & R(2+1)D-50 & Y8M & 67.4 & 93.8\\
    \midrule
    \textbf{Ours} & R(2+1)D-18 & Movies & 72.5 & 90.3\\
	\midrule
	\multicolumn{5}{c}{protocol: linear evaluation}\\
    \midrule
    STiCA\cite{Patrick_2021_ICCV} & R(2+1)D-18 & K400 & 48.2 & 77.0\\
    MMV\cite{alayrac2020self} & R(2+1)D-18 & AS & 60.0 & 83.9\\
    CVRL\cite{qian2021spatiotemporal} & R3D-50 & K400 & 57.3 & 89.2\\
    BraVe\cite{Recasens_2021_ICCV} & TSM-50 & AS & 69.1 & 93.4\\
    ELo\cite{piergiovanni2020evolving} & R(2+1)D-50 & Y8M & 64.5 & --\\
    \midrule
    \textbf{Ours} & R(2+1)D-18 & Movies & 64.6 & 80.8\\
    \midrule
	\multicolumn{5}{c}{protocol: linear Evaluation}\\
	\midrule
     Method & Arch. & Data & ESC50 & K400\\
     \midrule
    STiCA\cite{Patrick_2021_ICCV} & R(2+1)D-18 & K400 & 81.1 & --\\
    AVID\cite{morgado_avid_cma} & R(2+1)D-18 & K400 & 79.1 & 44.5\\
    XDC\cite{alwassel2019self} & R(2+1)D-18 & K400 & 78.5 & --\\
    AVID\cite{morgado_avid_cma} & R(2+1)D-18 & AS & 89.1 & 48.9\\
    MMV\cite{alayrac2020self} & R(2+1)D-18 & AS & 85.6 & --\\
    AVTS\cite{korbar2018cooperative} & MC3-18 & K400 & 76.7 & --\\
    AVTS\cite{korbar2018cooperative} & MC3-18 & AS & 80.6 & --\\
    CVRL\cite{qian2021spatiotemporal} & R3D-50 & K400 & -- & 66.1\\
    BraVe\cite{Recasens_2021_ICCV} & TSM-50 & AS & 92.1 & --\\
    \midrule
    \textbf{Ours} & R(2+1)D-18 & Movies & 83.6 & 43.6\\
    \bottomrule
  \end{tabular}
  \caption{Comparison with the state-of-the-art. Column ``Data'' indicates the pretraining dataset with abbreviations as follows: 
  \textbf{K}inetics-\textbf{400} \cite{kay2017kinetics}, \textbf{A}udio\textbf{S}et\cite{gemmeke2017audio}, \textbf{Y}outube-\textbf{8M}\cite{abu2016youtube}, \textbf{IG-}\textbf{K}inetics \cite{ghadiyaram2019large}, and \textbf{IG-}\textbf{R}andom \cite{alwassel2019self}. For HMDB51\cite{kuehne2011hmdb}, UCF101\cite{soomro2012ucf101}, and ESC50\cite{piczak2015esc}, we report the average results on all the folds.}
  \label{tab:sota}  
\end{table}

\subsection{Comparison with the state-of-the-art}\label{subsec:sota}

Table \ref{tab:sota} compares our proposed approach of learning from uncurated movies against the best performing audiovisual self-supervised learning methods. For the sake of fairness, we've included specifics of backbone architectures and pretraining datasets used in each method. In general, we achieve very competitive results on HMDB51\cite{kuehne2011hmdb}, however on UCF101\cite{soomro2012ucf101}, our numbers do fall behind. It is worth reminding that, the pretraining datasets used by other methods are all curated, with the exception of IG-Random\cite{alwassel2019self}, and are often significantly larger than our pretraining data. On ESC50\cite{piczak2015esc} and Kinetics-400\cite{kay2017kinetics}, we achieve comparable results with the state-of-the-art. For instance, on Kinetics-400\cite{kay2017kinetics}, while using the same backbone architecture, our model performs on par with AVID\cite{morgado_avid_cma} despite it has been pretrained on the same Kinetics-400\cite{kay2017kinetics} dataset. Finally, we did experiment with VGGSound\cite{Chen20} and obtained \textbf{0.38} and \textbf{0.48 mAP}, respectively in linear and finetuning evaluation regimes, though unfortunately none of the previously discussed self-supervised methods have reported on that benchmark.

\section{Conclusion}\label{sec:conclusion}
Despite its recent progress, state-of-the-art self-supervised learning literature has almost always relied on semantically curated datasets of short-form content for pretraining. Naturally, such a choice has influenced the direction of the proposed techniques, including the formulation of the objective function, and what constitutes as semantically similar/dissimilar. In this work, we studied self-supervised pretraining on semantically uncurated long-form content (\textit{i.e.} movies). We identified characteristics specific to movies, and comprehensively explored how within-content negative sampling can harness them to improve the quality of contrastive learning. Our experiments show that, pretraining on long-form content, even at a comparatively smaller scale to the curated and supervised alternatives, can give rise to representations capable of competing with the state-of-the-art. Keeping our approach and training strategy as simple as possible, we demonstrated that effectively learning audiovisual representations is feasible with less data and no human involvement. We emphasize that any machine learning method is susceptible to the potential underlying biases in the data. This is more important for self-supervised methods that deal with huge volumes, often not evaluated by diverse group of humans for any fairness concerns. The same is generally true in our case which requires us to make sure that titles that are included in the pretraining are diverse and inclusive.

\section*{Acknowledgments}
The authors are grateful to Dario Garcia-Garcia for the valuable discussions and feedback on the manuscript.

{\small
\bibliographystyle{ieee_fullname}
\bibliography{references}
}

\end{document}